\UseRawInputEncoding
\documentclass[10pt,journal,compsoc]{IEEEtran}
\usepackage{cite}
\usepackage{amsfonts}

\usepackage{verbatim}

\usepackage{multirow}
\ifCLASSINFOpdf
\usepackage[pdftex]{graphicx}
\DeclareGraphicsExtensions{.pdf,.jpeg,.png}

\else
\usepackage[dvips]{graphicx}
\usepackage[caption=false,font=footnotesize,labelfont=sf,textfont=sf]{subfig}
\graphicspath{{../eps/}}
\DeclareGraphicsExtensions{.eps}
\fi

\usepackage{amsmath}

\usepackage{algorithmic}
\usepackage{array}
\begin{document}
\title{Probabilistic K-means Clustering via\\ Nonlinear Programming}

\author{Yujian~Li*,
        Bowen~Liu,
        Zhaoying~Liu,
        and~Ting~Zhang
\IEEEcompsocitemizethanks{
\IEEEcompsocthanksitem Y. Li is with Faculty of
Information Technology, Beijing University of Technology, Beijing 100124, China. As corresponding author, he is with
School of Artificial Intelligence, Guilin University of Electronic Technology, Guilin 541004, China.\protect\\
E-mails: liyujian@guet.edu.cn
\IEEEcompsocthanksitem B. Liu, Z. Liu and T. Zhang are with Faculty of
Information Technology, Beijing University of Technology, Beijing 100124, China.\protect\\
E-mail: liubw2017@emails.bjut.edu.cn, zhaoying.liu@bjut.edu.cn,
zhangting@bjut.edu.cn}
\thanks{ }}

\markboth{}
{Shell \MakeLowercase{\textit{et al.}}: Probabilistic K-means Clustering via Nonlinear Programming}

\IEEEtitleabstractindextext{%
\begin{abstract}
K-means is a classical clustering algorithm with wide applications. However, soft K-means, or fuzzy c-means at \emph{m} = 1, remains unsolved since 1981. To address this challenging open problem, we propose a novel clustering model, i.e. Probabilistic K-Means (PKM), which is also a nonlinear programming model constrained on linear equalities and linear inequalities. In theory, we can solve the model by active gradient projection, while inefficiently. Thus, we further propose maximum-step active gradient projection and fast maximum-step active gradient projection to solve it more efficiently. By experiments, we evaluate the performance of PKM and how well the proposed methods solve it in five aspects: initialization robustness, clustering performance, descending stability, iteration number, and convergence speed.
\end{abstract}

}

\maketitle

\IEEEdisplaynontitleabstractindextext

\IEEEpeerreviewmaketitle

\IEEEraisesectionheading{\section{Introduction}\label{sec:introduction}}
\IEEEPARstart{C}{lustering} analysis is an unsupervised machine learning method. It has been widely used in image and video processing \cite{1}\! - \![4], speech processing \cite{5}, biology \cite{6}, medicine \cite{7}, sociology \cite{8}, and so on. The task of clustering is to find some similarities from datasets \cite{9}, \cite{10}, and then to classify samples with the similarities into clusters (i.e. classes or categories). Clustering methods mainly include: partition-based clustering \cite{11}\! -\!\cite{13}, hierarchical-based clustering \cite{14}\! -\!\cite{16}, density-based clustering \cite{10},\cite{17},\cite{18}, graph-based clustering \cite{19}\! - \!\cite{21}, grid-based clustering \cite{22}\! -\!\cite{24}, model-based clustering \cite{25}\! -\!\cite{27} and subspace clustering \cite{21}, \cite{24}, \cite{28}\! -\!\cite{31}, etc. Note that these methods may have intersections.

\par Partition-based clustering algorithms, such as K-means \cite{11}, K-medoids \cite{12}, and EM-K-means \cite{32}, divide a dataset into several disjoint subsets with a similarity criterion. In general, they first choose initial cluster centers randomly or manually, then adjust categories of samples and update the cluster centers according to the nearest neighbor principle until convergence. Hierarchical-based clustering algorithms fall into two types: bottom-up and top-down. The bottom-up clustering algorithms, such as CURE \cite{14} and BIRCH \cite{15}, start with each sample treated as a separate cluster, and repeatedly merge two or more clusters until satisfying certain conditions, e.g. only one cluster left. The top-down clustering algorithms, like MMDCA \cite{16}, start with a single cluster of all samples, and repeatedly split one cluster into two or more dissimilar clusters until meeting some criteria, e.g. only one sample in each cluster. Density-based clustering algorithms, such as DBSCAN \cite{17}, "affinity propagation" \cite{10} and "density peak" \cite{18}, cluster samples according to their density distributions with local centers, where the definition of density may change in different algorithms.
\par Graph-based clustering algorithms include variational clustering \cite{19}, minimum spanning tree \cite{20} and spectral clustering \cite{29}, etc. These algorithms construct a similarity graph for a dataset and then find an optimal partition such that the graph has high weights for edges within each group and low weights for edges between different groups. Grid-based clustering algorithms, such as STING \cite{22}, WaveCluster \cite{23} and CLIQUE \cite{24}, classify datasets with spatial grid cells, which may contain different numbers of data points. Model-based clustering algorithms mainly include self-organizing map (SOM) \cite{25} and statistical clustering \cite{26}. SOM is an unsupervised neural network for projecting all data points onto a two-dimensional plane with similarities between them. Statistical clustering classifies datasets with their probability distributions in optimization of expectation maximization \cite{27}.
\par Subspace clustering algorithms find several subspaces from a data set, where each subspace contains only one cluster. These clustering algorithms may also contain grid-based CLIQUE \cite{24}, model-based RAS \cite{26}, and graph-based spectral clustering \cite{28}\! -\!\cite{31}, etc.
\par K-means clustering algorithms minimize the sum of squared (Euclidean) distances between each sample point and its nearest cluster center, mainly including Lloyd's algorithm \cite{33}, Hartigan-Wong's algorithm \cite{34} and MacQueen's algorithm \cite{11}. These heuristic algorithms take advantages of easy implementation, high efficiency and low complexity. However, they are sensitive to initial centers and outliers, and may converge to partitions that are significantly inferior to the global optimum. Meanwhile, they are not suitable for finding non-spherical clusters in a dataset. To improve them, many new algorithms have been proposed, such as GA-K-means, K-means++, K-medoids, $L_{p}$-norm K-means, symmetric distance K-means, and kernel K-means. GA-K-means \cite{35} uses a genetic algorithm to select initial centers, and can significantly improve clustering performance on low-dimensional datasets. K-means++ \cite{36} sets initial centers to be far apart data points, and can improve the stability of clustering results. K-medoids \cite{12} replaces mean centers with some special samples called medoids, and can reduce the sensitivity to outliers. $L_{p}$-norm K-means \cite{37} exploits Minkowski distance to measure similarities between samples, and can be thought of as an extension of K-means. Symmetric distance K-means \cite{38} employs a non-metric "point symmetric" distance for clustering, and can find different-shape clusters. Kernel K-means \cite{39} selects a kernel function to cluster in a feature space, and can find non-linear separable structures. Ref. \cite{56} extends K-means type algorithms  by integrating both intracluster compactness and intercluster separation.
\par Another well-known variant of K-means is fuzzy c-means (FCM) \cite{13}  proposed by Bezdek in 1981. FCM determines partitions by computing the membership degree to each cluster for each data point. The higher the membership degree of a data point to a cluster, the greater possibility the data point belongs to the cluster. Compared with K-means, FCM is more flexible in applications, but with one more parameter \emph{m} to be selected manually. In theory, \emph{m} ranges from 1 to infinity. In practice, \emph{m} $\approx$ 1.3 is a relatively good value for clustering \cite{40}. As \emph{m} approaches 1, FCM degenerates to K-means, with all membership degrees converging to 0 or 1. Additionally, FCM has many variants, such as possibilistic c-means \cite{41}, sparse possibilistic c-means \cite{42}, and size insensitive integrity-based FCM \cite{43}. However, a remaining problem is how to solve FCM at \emph{m} = 1. The problem will be solved by a novel clustering model, called probabilistic K-means (PKM).
\par In fact, PKM is an equivalent model to FCM at \emph{m} = 1. Theoretically, PKM is also a nonlinear programming problem with many constraints of linear equalities and linear inequalities. It can be solved by an active gradient projection (AGP) method, i.e. a combination of active set \cite{44} and gradient projection \cite{45}, \cite{46}. To speed up the AGP method, we further present a maximum-step AGP (MSAGP) method by iteratively computing maximum feasible step length and a fast MSAGP (FMSAGP) method by efficiently computing projection matrices. Also, we evaluate the performance of PKM and how well the proposed methods solve it on artificial and real datasets.
\par The remainder of this paper is organized as follows: Section 2 introduces K-means, fuzzy c-means and probabilistic K-means. Section 3 presents solutions to PKM in detail. Experimental results are reported in Section 4. Finally, conclusions are drawn in Section 5.




\section{PKM Clustering Model\label{PKM clusteing model}}
\par Given a set of data points ${\textbf{\emph{X}}} = \left\{ {{{\mathbf{\textbf{\emph{x}}}}_i}|{{\mathbf{\textbf{\emph{x}}}}_i} \in {\mathbb{R}^D},1 \le i \le L} \right\}$, let us divide it into \emph{K} clusters. Suppose ${\omega _j}$  is the \emph{j}-th cluster. Then, ${\mathbf{\emph{\textbf{X}}}} = \bigcup\limits_{j = 1}^K {{\omega _j}}$, $\;\forall 1 \leq i \ne j \leq K, { \omega _i}\bigcap {{\omega _j} = \emptyset } $. If ${L_j}$ denotes the number of elements in ${\omega _j}$ with the center of ${{\textbf{\emph{c}}}_j}$, and $L = \sum\limits_{i = 1}^K {{L_j}} $, then the standard K-means clustering model is described as minimizing a ¡°hard¡± objective function:
\begin{equation}\label{eq1}
J = \sum\limits_{i = 1}^K {\sum\limits_{{{\mathbf{\emph{\textbf{x}}}}_i} \in {\omega _j}} {{{\left\| {{{\mathbf{\emph{\textbf{x}}}}_i} - {{\mathbf{\emph{\textbf{c}}}}_j}} \right\|}^2}} }
\end{equation}
 where
\begin{equation}\label{eq2}
{{\mathbf{\emph{\textbf{c}}}}_j} = \frac{1}{{{L_j}}}\sum\limits_{{{\mathbf{\textbf{\emph{x}}}}_i} \in {\omega _j}} {{{\mathbf{\textbf{\emph{x}}}}_i}}
\end{equation}
\par With membership degree ${w_{ij}}$ assigned to the \emph{i}-th data point in the \emph{j}-th cluster for $1 \leq i \leq L$ and $1 \leq j \leq K$, the ¡°hard¡± K-means model is extended to a soft model, namely the FCM model. FCM minimizes the objective function below: 
\begin{equation}\label{eq3}
\begin{gathered}
  J = \sum\limits_{j = 1}^K {\sum\limits_{i = 1}^L {w_{_{ij}}^m{{\left\| {{{\mathbf{\textbf{\emph{x}}}}_i} - {{\mathbf{\textbf{\emph{c}}}}_j}} \right\|}^2}} }  \hfill \\
  s.t.\sum\limits_{j = 1}^K {{w_{ij}} = 1} ,{w_{ij}} \geq 0 \hfill \\
\end{gathered}
\end{equation}
where \emph{m} ranges from 1 to infinity.
\par When \emph{m} $>$ 1, ${w_{ij}}$ and ${\emph{\textbf{c}}_{j}}$ can be computed alternately as follows \cite{13}
\begin{equation}\label{eq4}
\left\{ \begin{gathered}
  {w_{ij}} = \frac{{{{\left\| {{{\mathbf{\textbf{\emph{x}}}}_i} - {{\mathbf{\textbf{\emph{c}}}}_j}} \right\|}^{ - \frac{2}{{m - 1}}}}}}{{\sum\limits_{k = 1}^K {{{\left\| {{{\mathbf{\textbf{\emph{x}}}}_i} - {{\mathbf{\textbf{\emph{c}}}}_k}} \right\|}^{ - \frac{2}{{m - 1}}}}} }} \hfill \\
  {{\mathbf{\textbf{\emph{c}}}}_j} = \frac{{\sum\limits_{i = 1}^L {w_{_{ij}}^m{{\mathbf{\textbf{\emph{x}}}}_i}} }}{{\sum\limits_{i = 1}^L {w_{_{ij}}^m} }} \hfill \\
\end{gathered}  \right.
\end{equation}
\par However, FCM at \emph{m} = 1 is an challenging open problem that remains unsolved since 1981. It is also called "soft K-means". Replacing  membership degree ${w_{ij}}$ by probability  ${p_{ij}}$, we get the soft K-means clustering model as follows,
\begin{equation}\label{eq5}
\begin{gathered}
  J = \sum\limits_{j = 1}^K {\sum\limits_{i = 1}^L {{p_{ij}}{{\left\| {{{\mathbf{\textbf{\emph{x}}}}_i} - {{\mathbf{\textbf{\emph{c}}}}_j}} \right\|}^2}} }  \hfill \\
  s.t.\sum\limits_{j = 1}^K {{p_{ij}} = 1} ,{p_{ij}} \geq 0 \hfill \\
\end{gathered}
\end{equation}
\par Note that probability ${p_{ij}}$      takes a value in [0, 1]. This means that data point ${\emph{\textbf{x}}_{i}}$  may belong to class \emph{j} with probability  ${p_{ij}}$, rather than to only one class in hard K-means. The higher the probability ${p_{ij}}$, the more likely that ${\emph{\textbf{x}}_{i}}$  is in class \emph{j}. The view of probability is more in line with Bayesian decision theory \cite{47}.
\par Additionally, we have to emphasize that the soft K-means, namely (\ref{eq5}), is not equivalent to the optimization problem of hard K-means, namely, (\ref{eq1}) and (\ref{eq2}). On the one hand, ${{\textbf{\emph{c}}}_j}$ may have something to do with $p_{ij}$. On the other hand, if setting $p_{ij}=1$  with the minimal value of $\left\| {{{\textbf{\emph{x}}}_i} - {{\textbf{\emph{c}}}_j}} \right\|$  for $1\leq j\leq K$ and ${p_{ik}} = 0$  for $k\neq j$, we cannot compute ${{\textbf{\emph{c}}}_j}$  as (\ref{eq2}) in hard K-means because of no definition for ${{\textbf{\emph{c}}}_j}$ in (\ref{eq5}).
\par For the soft K-means, we can construct a Lagrangian function,
\begin{equation}\label{eq6}
J = \sum\limits_{j = 1}^K {\sum\limits_{i = 1}^L {{p_{ij}}{{\left\| {{{\mathbf{\textbf{\emph{x}}}}_i} - {{\mathbf{\textbf{\emph{c}}}}_j}} \right\|}^2}} }  + \sum\limits_{i = 1}^L {{\lambda _i}(\sum\limits_{j = 1}^K {{p_{ij}} - 1} )}
\end{equation}
\par Letting $\frac{{\partial J}}{{\partial {\textbf{\emph{c}}_j}}} = \sum\limits_{j = 1}^K {\sum\limits_{i = 1}^L {2{p_{ij}}({{\mathbf{\textbf{\emph{c}}}}_j} - {{\mathbf{\textbf{\emph{x}}}}_i}) = 0} } $, we obtain
\begin{equation}\label{eq7}
{{\mathbf{\textbf{\emph{c}}}}_j} = \frac{{\sum\limits_{i = 1}^L {{p_{ij}}{{\mathbf{\textbf{\emph{x}}}}_i}} }}{{\sum\limits_{i = 1}^L {{p_{ij}}} }}
\end{equation}
\par Substituting (\ref{eq7}) into (\ref{eq5}), we get an equivalent model, namely, PKM
\begin{equation}\label{eq8}
\begin{gathered}
  J = \sum\limits_{j = 1}^K {\sum\limits_{i = 1}^L {{p_{ij}}{{\left\| {{{\mathbf{\textbf{\emph{x}}}}_i} - \frac{{\sum\limits_{i = 1}^L {{p_{ij}}{{\mathbf{\textbf{\emph{x}}}}_i}} }}{{\sum\limits_{i = 1}^L {{p_{ij}}} }}} \right\|}^2}} }  \hfill \\
  s.t.\sum\limits_{j = 1}^K {{p_{ij}} = 1} ,{p_{ij}} \geq 0 \hfill \\
\end{gathered}
\end{equation}
\par If setting \emph{K} = 2 and \emph{L} = 2  with ${{\mathbf{\textbf{\emph{x}}}}_1} = {(1,1)^{\text{T}}},\text{ } {{\mathbf{\textbf{\emph{x}}}}_2} = {(2,2)^{\text{T}}}$, we can rewrite the objective function as follows,
\begin{equation}\label{eq9}
\begin{gathered}
  J({p_{11}},{p_{12}},{p_{21}},{p_{22}}) = \sum\limits_{j = 1}^2 {\sum\limits_{i = 1}^2 {{p_{ij}}{{\left\| {{{\mathbf{\textbf{\emph{x}}}}_i} - \frac{{\sum\limits_{i = 1}^2 {{p_{ij}}{{\mathbf{\textbf{\emph{x}}}}_i}} }}{{\sum\limits_{i
  = 1}^2 {{p_{ij}}} }}} \right\|}^2}} }  \hfill \\
   = \left[ {{p_{11}}{{\left\| {\frac{{{p_{21}}{{\mathbf{\textbf{\emph{x}}}}_1} - {p_{21}}{{\mathbf{\textbf{\emph{x}}}}_2}}}{{{p_{11}} + {p_{21}}}}} \right\|}^2} + {p_{21}}{{\left\| {\frac{{{p_{11}}{{\mathbf{\textbf{\emph{x}}}}_2} - {p_{11}}{{\mathbf{\textbf{\emph{x}}}}_1}}}{{{p_{11}} + {p_{21}}}}} \right\|}^2}} \right] \hfill \\
   + \left[ {{p_{12}}{{\left\| {\frac{{{p_{22}}{{\mathbf{\textbf{\emph{x}}}}_1} - {p_{22}}{{\mathbf{\textbf{\emph{x}}}}_2}}}{{{p_{12}} + {p_{22}}}}} \right\|}^2} + {p_{22}}{{\left\| {\frac{{{p_{12}}{{\mathbf{\textbf{\emph{x}}}}_2} - {p_{12}}{{\mathbf{\textbf{\emph{x}}}}_1}}}{{{p_{12}} + {p_{22}}}}} \right\|}^2}} \right]  \hfill \\
   = \frac{{{p_{11}}{p_{12}} + {p_{21}}{p_{22}}}}{{\left( {{p_{11}} + {p_{21}}} \right)\left( {{p_{12}} + {p_{22}}} \right)}}{\left\| {{{\mathbf{\textbf{\emph{x}}}}_1} - {{\mathbf{\textbf{\emph{x}}}}_2}} \right\|^2} \hfill \\
   = \frac{{{p_{11}}\left( {1 - {p_{11}}} \right) + {p_{21}}\left( {1 - {p_{21}}} \right)}}{{\left( {{p_{11}} + {p_{21}}} \right)\left( {2 - {p_{11}} - {p_{21}}} \right)}}{\left\| {{{\mathbf{\textbf{\emph{x}}}}_1} - {{\mathbf{\textbf{\emph{x}}}}_2}} \right\|^2} \hfill \\
   = \frac{{{p_{11}} + {p_{21}} - p_{11}^2 - p_{21}^2}}{{\left( {{p_{11}} + {p_{21}}} \right)\left( {2 - {p_{11}} - {p_{21}}} \right)}}{\left\| {{{\mathbf{\textbf{\emph{x}}}}_1} - {{\mathbf{\textbf{\emph{x}}}}_2}} \right\|^2}
   = J({p_{11}},{p_{21}}),
\end{gathered}
\end{equation}
where  ${p_{12}} = 1 - {p_{11}}$ $\left( {0 \leq {p_{11}} \leq 1} \right)$ and ${p_{22}} = 1 - {p_{21}} $ $\left( {0 \leq {p_{21}} \leq 1} \right)$.

\begin{figure}[htb]
\centering
\includegraphics[width=0.2\textwidth]{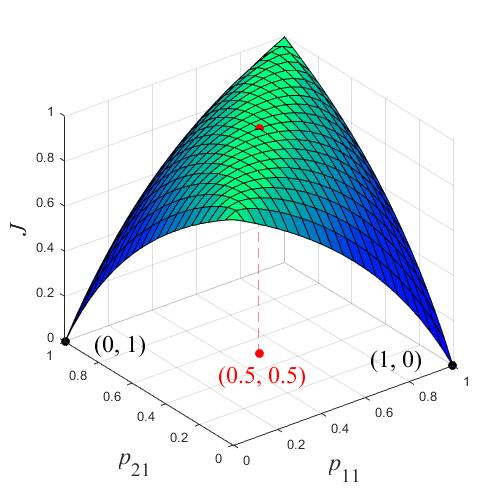} 
\includegraphics[width=0.2\textwidth]{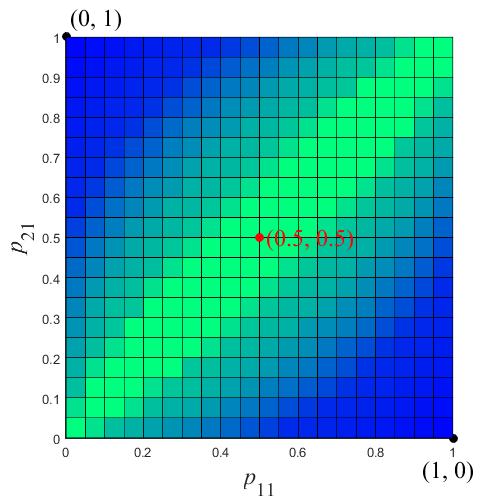} \\
(a)\texttt{ }\texttt{ }\texttt{ }\texttt{ }\texttt{ }\texttt{ }\texttt{ }\texttt{ }\texttt{ }\texttt{ }\texttt{ }\texttt{ }\texttt{ }\texttt{ }\texttt{ }\texttt{ } \texttt{ }(b)
\caption{A surface of $J({p_{11}},{p_{21}})$  (a), and its vertical view (b).}
\label{fig_1}
\end{figure}

\par Therefore, $J({p_{11}},{p_{12}},{p_{21}},{p_{22}}) = J({p_{11}},{p_{21}})$  is indeed a function of $p_{11}$ and $p_{21}$, as shown in Fig. \ref{fig_1}. Note that Fig. \ref{fig_1}a is a surface of $J({p_{11}},{p_{21}})$, with its vertical view in Fig. \ref{fig_1}b. In Fig. \ref{fig_1}, $p_{11}$ is the probability of $\textbf{\emph{x}}_{1}$ in cluster 1, and $p_{21}$ is the probability of $\textbf{\emph{x}}_{2}$ in cluster 1. Obviously, the function $J({p_{11}},{p_{21}})$  has countless probability pairs that take the same maximum value of 1, satisfying ${p_{11}} = {p_{21}}$, e.g. (0.5, 0.5). But it has only two probability pairs that take the same minimum value of 0, namely (0, 1) or (1, 0). Because of symmetricity, the two pairs produce the same cluster assignation: $\textbf{\emph{x}}_{1}$ in one class, and $\textbf{\emph{x}}_{2}$ in the other.

\section{Solutions to PKM}\label{Solutions to PKM}
According to (\ref{eq8}), PKM is a nonlinear programming problem constrained on linear equalities and linear inequalities. In theory, the problem can be solved by AGP, but requiring optimization of step length and fast computation of projection matrices.
\par In this section, we first calculate the gradient of PKM's objective function, and then propose the AGP method and its two improvements: the MSAGP method and the FMSAGP method, to solve the PKM model.

\subsection{Gradient Calculation}\label{Gradient Calculation}
For convenience, we define ${F_k}$  as
\begin{equation}\label{eq10}
{F_k} = {\left\| {{{\mathbf{\textbf{\emph{x}}}}_k} - \frac{{\sum\limits_{i = 1}^L {{p_{ij}}{{\mathbf{\textbf{\emph{x}}}}_i}} }}{{\sum\limits_{i = 1}^L {{p_{ij}}} }}} \right\|^2}
\end{equation}
\par Using the chain rule on (\ref{eq8}), we obtain
\begin{equation}\label{eq11}
\frac{{\partial J}}{{\partial {p_{ij}}}} ={F_i} +  \sum\limits_{k = 1}^L {{p_{kj}}\frac{{\partial {F_k}}}{{\partial {p_{ij}}}}}
\end{equation}
According to (\ref{eq10}), we can further derive (\ref{eq12})-(\ref{eq15}) as follows,
\begin{equation}\label{eq12}
\begin{gathered}
  \frac{{\partial {F_k}}}{{\partial {p_{ij}}}} = \frac{\partial }{{\partial {p_{ij}}}}\left[ {{{\mathbf{\emph{\textbf{x}}}}_k}^{\text{T}}{{\mathbf{\textbf{\emph{x}}}}_k} - {{\mathbf{\emph{\textbf{x}}}}_k}^{\text{T}}\left( {\frac{{\sum\limits_{i = 1}^L {{p_{ij}}{{\mathbf{\textbf{\emph{x}}}}_i}} }}{{\sum\limits_{i = 1}^L {{p_{ij}}} }}} \right)} \right. \hfill \\
  \left. { - {{\left( {\frac{{\sum\limits_{i = 1}^L {{p_{ij}}{{\mathbf{\emph{\textbf{x}}}}_i}} }}{{\sum\limits_{i = 1}^L {{p_{ij}}} }}} \right)}^{\text{T}}}{{\mathbf{\textbf{\emph{x}}}}_k} + {{\left( {\frac{{\sum\limits_{i = 1}^L {{p_{ij}}{{\mathbf{\textbf{\emph{x}}}}_i}} }}{{\sum\limits_{i = 1}^L {{p_{ij}}} }}} \right)}^{\text{T}}}\left( {\frac{{\sum\limits_{i = 1}^L {{p_{ij}}{{\mathbf{\textbf{\emph{x}}}}_i}} }}{{\sum\limits_{i = 1}^L {{p_{ij}}} }}} \right)} \right] \hfill \\
\end{gathered}
\end{equation}

\begin{equation}\label{eq13}
\begin{gathered}
  \frac{{\partial {F_k}}}{{\partial {p_{ij}}}} =  - 2\frac{{{\mathbf{\emph{\textbf{x}}}}_k^{\text{T}}{{\mathbf{\emph{\textbf{x}}}}_i}\sum\limits_{i = 1}^L {{p_{ij}}}  - {\mathbf{\textbf{\emph{x}}}}_k^{\text{T}}\sum\limits_{i = 1}^L {{p_{ij}}{{\mathbf{\textbf{\emph{x}}}}_i}} }}{{{{\left( {\sum\limits_{i = 1}^L {{p_{ij}}} } \right)}^2}}} \hfill \\
   + 2\frac{{{\mathbf{\textbf{\emph{x}}}}_i^{\text{T}}\left( {\sum\limits_{i = 1}^L {{p_{ij}}{{\mathbf{\textbf{\emph{x}}}}_i}} } \right)\sum\limits_{i = 1}^L {{p_{ij}}}  - \left[ {\left( {\sum\limits_{i = 1}^L {{p_{ij}}{\mathbf{\textbf{\emph{x}}}}_i^{\text{T}}} } \right)\sum\limits_{i = 1}^L {{p_{ij}}{{\mathbf{\textbf{\emph{x}}}}_i}} } \right]}}{{{{\left( {\sum\limits_{i = 1}^L {{p_{ij}}} } \right)}^3}}} \hfill \\
\end{gathered}
\end{equation}

\begin{equation}\label{eq14}
\begin{gathered}
  \frac{{\partial {F_k}}}{{\partial {p_{ij}}}} = \frac{{ - 2}}{{\sum\limits_{i = 1}^L {{p_{ij}}} }}\left( {{\mathbf{\emph{\textbf{x}}}}_k^{\text{T}}{{\mathbf{\textbf{\emph{x}}}}_i} - \frac{{{\mathbf{\textbf{\emph{x}}}}_k^{\text{T}}\sum\limits_{i = 1}^L {{p_{ij}}{{\mathbf{\emph{\textbf{x}}}}_i}} }}{{\sum\limits_{i = 1}^L {{p_{ij}}} }}} \right. \hfill \\
  \left. { - \frac{{{\mathbf{\textbf{\emph{x}}}}_i^{\text{T}}\left( {\sum\limits_{i = 1}^L {{p_{ij}}{{\mathbf{\textbf{\emph{x}}}}_i}} } \right)}}{{\sum\limits_{i = 1}^L {{p_{ij}}} }} + \frac{{\left[ {\left( {\sum\limits_{i = 1}^L {{p_{ij}}{\mathbf{\textbf{\emph{x}}}}_i^{\text{T}}} } \right)\sum\limits_{i = 1}^L {{p_{ij}}{{\mathbf{\textbf{\emph{x}}}}_i}} } \right]}}{{{{\left( {\sum\limits_{i = 1}^L {{p_{ij}}} } \right)}^2}}}} \right) \hfill \\
\end{gathered}
\end{equation}
\begin{equation}\label{eq15}
\frac{{\partial {F_k}}}{{\partial {p_{ij}}}} =  - \frac{2}{{\sum\limits_{i = 1}^L {{p_{ij}}} }}{\left( {{{\mathbf{\textbf{\emph{x}}}}_k} - \frac{{\sum\limits_{i = 1}^L {{p_{ij}}{{\mathbf{\textbf{\emph{x}}}}_i}} }}{{\sum\limits_{i = 1}^L {{p_{ij}}} }}} \right)^{\text{T}}}\left( {{{\mathbf{\textbf{\emph{x}}}}_i} - \frac{{\sum\limits_{i = 1}^L {{p_{ij}}{{\mathbf{\textbf{\emph{x}}}}_i}} }}{{\sum\limits_{i = 1}^L {{p_{ij}}} }}} \right)
\end{equation}
Substituting (\ref{eq15}) into (\ref{eq11}), we finally get,
\begin{equation}\label{eq16}
\begin{gathered}
  \frac{{\partial J}}{{\partial {p_{ij}}}} = {\left\| {{{\mathbf{\textbf{\emph{x}}}}_i} - \frac{{\sum\limits_{i = 1}^L {{p_{ij}}{{\mathbf{\textbf{\emph{x}}}}_i}} }}{{\sum\limits_{i = 1}^L {{p_{ij}}} }}} \right\|^2} \hfill \\
   - \frac{2}{{\sum\limits_{i = 1}^L {{p_{ij}}} }}{\sum\limits_{k = 1}^L {{p_{kj}}\left( {{{\mathbf{\textbf{\emph{x}}}}_k} - \frac{{\sum\limits_{i = 1}^L {{p_{ij}}{{\mathbf{\textbf{\emph{x}}}}_i}} }}{{\sum\limits_{i = 1}^L {{p_{ij}}} }}} \right)} ^{\text{T}}}\left( {{{\mathbf{\textbf{\emph{x}}}}_i} - \frac{{\sum\limits_{i = 1}^L {{p_{ij}}{{\mathbf{\textbf{\emph{x}}}}_i}} }}{{\sum\limits_{i = 1}^L {{p_{ij}}} }}} \right) \hfill \\
\end{gathered}
\end{equation}
\begin{equation}\label{eq17}
\begin{gathered}
  \frac{{\partial J}}{{\partial {p_{ij}}}} = \left[ {\left( {{{\mathbf{\textbf{\emph{x}}}}_i} - \frac{{\sum\limits_{i = 1}^L {{p_{ij}}{{\textbf{\emph{x}}}_i}} }}{{\sum\limits_{i = 1}^L {{p_{ij}}} }}} \right)} \right. \hfill \\
  {\left. { - \frac{2}{{\sum\limits_{i = 1}^L {{p_{ij}}} }}\sum\limits_{k = 1}^L {{p_{kj}}\left( {{{\textbf{\emph{x}}}_k} - \frac{{\sum\limits_{i = 1}^L {{p_{ij}}{{\textbf{\emph{x}}}_i}} }}{{\sum\limits_{i = 1}^L {{p_{ij}}} }}} \right)} } \right]^{\text{T}}}\left( {{{\textbf{\emph{x}}}_i} - \frac{{\sum\limits_{i = 1}^L {{p_{ij}}{{\textbf{\emph{x}}}_i}} }}{{\sum\limits_{i = 1}^L {{p_{ij}}} }}} \right) \hfill \\
\end{gathered}
\end{equation}
and the gradient

\begin{equation}\label{eq18}
\begin{gathered}
  \nabla J = \left[ {\begin{array}{*{20}{c}}
  {\frac{{\partial J}}{{\partial {p_{11}}}}}&{...}&{\frac{{\partial J}}{{\partial {p_{1K}}}}}&{...}
\end{array}} \right. \hfill \\
  {\text{ }}\left. {\begin{array}{*{20}{c}}
  {\frac{{\partial J}}{{\partial {p_{ij}}}}}&{...}&{\frac{{\partial J}}{{\partial {p_{Lj}}}}}&{...}&{\frac{{\partial J}}{{\partial {p_{L\!K}}}}}
\end{array}} \right] \hfill \\
\end{gathered}
\end{equation}

\subsection{The AGP Method}\label{The AGP Method}
In the constraints of PKM, there are \emph{L} equalities and \emph{KL} inequalities,
\begin{equation}\label{eq19}
\forall 1 \leq i \leq L,\sum\limits_{j = 1}^K {{p_{ij}} = 1}
\end{equation}
\begin{equation}\label{eq20}
\forall 1 \leq i \leq L,1 \leq j \leq K,{p_{ij}} \geq 0
\end{equation}
where probability matrix ${\left( {{p_{ij}}} \right)_{L \times K}}$  can be vectorized as
\begin{equation}\label{eq21}
{\mathbf{\textbf{\emph{P}}}} = {\left[ {\begin{array}{*{20}{c}}
  {{p_{11}}}&{...}&{{p_{1K}}}&{...}&{\begin{array}{*{20}{c}}
  {{p_{ij}}}&{...}&{\begin{array}{*{20}{c}}
  {{p_{Lj}}}&{...}&{{p_{L\!K}}}
\end{array}}
\end{array}}
\end{array}} \right]^{\text{T}}}
\end{equation}
\par Let ${{\mathbf{\textbf{\emph{I}}}}_{LK \times LK}}$  be the identity matrix of size $LK\times LK$. Define two matrices, inequality matrix \textbf{\emph{A}} and equality matrix \textbf{\emph{E}},
\begin{equation}\label{eq22}
{\mathbf{\textbf{\emph{A}}}} = {{\mathbf{\textbf{\emph{I}}}}_{LK \times LK}}
\end{equation}

\begin{equation}\label{eq23}
{\mathbf{\emph{\textbf{E}}}} = {\left[ {\begin{array}{*{20}{c}}
  {\underbrace {\begin{array}{*{20}{c}}
  1&{...}&1
\end{array}}_K}&{\begin{array}{*{20}{c}}
  0&0&{...}
\end{array}}&{\begin{array}{*{20}{c}}
  0&0&0
\end{array}} \\
  {\underbrace {\begin{array}{*{20}{c}}
  0&{...}&0
\end{array}}_K}&{\underbrace {\begin{array}{*{20}{c}}
  1&{...}&1
\end{array}}_K}&{\begin{array}{*{20}{c}}
  0&{...}&0
\end{array}} \\
  {}&{...}&{} \\
  {\begin{array}{*{20}{c}}
  0&0&{...}
\end{array}}&{\begin{array}{*{20}{c}}
  0&0&0
\end{array}}&{\underbrace {\begin{array}{*{20}{c}}
  1&{...}&1
\end{array}}_K}
\end{array}} \right]_{L \times LK}}
\end{equation}

\par Obviously, each row of \textbf{\emph{A}} corresponds to one and only one inequality in (\ref{eq20}). Moreover, each row of \textbf{\emph{E}} corresponds to one and only one equality in (\ref{eq19}). Accordingly, the PKM's constraints, namely, (\ref{eq19})-(\ref{eq20}), can be simply expressed as
\begin{equation}\label{eq24}
{\mathbf{\textbf{\emph{AP}}}} \geq {\textbf{0}},{\mathbf{\textbf{\emph{EP}}}} = {\textbf{1}}
\end{equation}
\par Let ${{\mathbf{\textbf{\emph{P}}}}^{(0)}}$  stand for an initial probability vector, and ${{\mathbf{\textbf{\emph{P}}}}^{(k)}}$  for the probability vector at iteration \emph{k}. Note that  ${{\mathbf{\textbf{\emph{P}}}}^{(0)}}$ is randomly initialized with the constraints of (\ref{eq24}). At iteration \emph{k}, the rows of inequality matrix \textbf{\emph{A}} are broken into two groups: one is active, the other is inactive. The active group is composed of all inequalities that must work exactly as an equality at ${{\mathbf{\textbf{\emph{P}}}}^{(k)}}$, whereas the inactive group is composed of the left inequalities. If ${\mathbf{\textbf{\emph{A}}}}_1^{(k)}$  and ${\mathbf{\textbf{\emph{A}}}}_2^{(k)}$  respectively denote the active group and the inactive group, we have
\begin{equation}\label{eq25}
{\mathbf{\textbf{\emph{A}}}}_1^{(k)}{{\mathbf{\textbf{\emph{P}}}}^{(k)}} = {\textbf{0}}
\end{equation}
\begin{equation}\label{eq26}
{\mathbf{\textbf{\emph{A}}}}_2^{(k)}{{\mathbf{\textbf{\emph{P}}}}^{(k)}} > {\textbf{0}}
\end{equation}

Combining ${\mathbf{\textbf{\emph{A}}}}_1^{(k)}$  with \textbf{\emph{E}}, we define an active matrix  ${{\mathbf{\textbf{\emph{N}}}}^{(k)}}$, namely,
\begin{equation}\label{eq27}
{{\mathbf{\textbf{\emph{N}}}}^{(k)}} = \left[ \begin{gathered}
  {\mathbf{\textbf{\emph{A}}}}_1^{(k)} \hfill \\
  {\mathbf{\textbf{\emph{E}}}} \hfill \\
\end{gathered}  \right]
\end{equation}
which determines the active constraints at iteration \emph{k}.
\par When ${\mathbf{\emph{\textbf{N}}}} = {{\mathbf{\textbf{\emph{N}}}}^{(k)}}$ is not a square matrix, we construct two matrices ${{\mathbf{\emph{\textbf{G}}}}^{(k)}}$  and ${{\mathbf{\textbf{\emph{Q}}}}^{(k)}}$  as follows,
\begin{equation}\label{eq28}
{{\mathbf{\textbf{\emph{G}}}}^{(k)}} = {{\mathbf{\textbf{\emph{N}}}}^{\text{T}}}{({\mathbf{\textbf{\emph{N}}}}{{\mathbf{\textbf{\emph{N}}}}^\text{T}})^{ - 1}}{\mathbf{\textbf{\emph{N}}}}
\end{equation}
\begin{equation}\label{eq29}
{{\mathbf{\textbf{\emph{Q}}}}^{(k)}} = {{\mathbf{\textbf{\emph{I}}}}_{LK \times LK}} - {{\mathbf{\textbf{\emph{G}}}}^{(k)}}
\end{equation}
where ${{\mathbf{\emph{\textbf{G}}}}^{(k)}}$   is called projection matrix, and ${{\mathbf{\textbf{\emph{Q}}}}^{(k)}}$  is its orthogonal projection matrix \cite{45}.
\par At iteration \emph{k}, projecting the gradient of PKM's objective function on the subspace of the active constraints, we obtain the projected gradient,
\begin{equation}\label{eq30}
{{\mathbf{\textbf{\emph{d}}}}^{(k)}} =  - {{\mathbf{\textbf{\emph{Q}}}}^{(k)}}\nabla J({{\mathbf{\textbf{\emph{P}}}}^{(k)}})
\end{equation}
\par If ${{\mathbf{\textbf{\emph{d}}}}^{(k)}} = {\textbf{0}}$, we stop at a local minimum. Otherwise, we compute the probability vector at iteration \emph{k} + 1,
\begin{equation}\label{eq31}
{{\mathbf{\textbf{\emph{P}}}}^{\left( {k + 1} \right)}} = {{\mathbf{\textbf{\emph{P}}}}^{\left( k \right)}} + t{{\mathbf{\textbf{\emph{d}}}}^{(k)}}
\end{equation}
where \emph{t} is a step length. Usually, \emph{t} takes a small value, e.g. $t = 0.01$  or $t = 0.1$.
\par When ${\mathbf{\textbf{\emph{N}}}} = {{\mathbf{\textbf{\emph{N}}}}^{(k)}}$ is a square matrix, \textbf{\emph{N}} must be invertible. Meanwhile, the projection matrix ${{{\textbf{\emph{G}}}}^{(k)}} = {{{\textbf{\emph{N}}}}^{\text{T}}}{({\mathbf{\textbf{\emph{N}}}}{{\mathbf{\textbf{\emph{N}}}}^\text{T}})^{ - 1}}{\mathbf{\textbf{\emph{N}}}} = {{\mathbf{\textbf{\emph{I}}}}_{LK \times LK}}$, and ${{\mathbf{\textbf{\emph{Q}}}}^{(k)}} = {\textbf{0}}$. Thus, the projected gradient  ${{\mathbf{\textbf{\emph{d}}}}^{(k)}} =  - {{\mathbf{\textbf{\emph{Q}}}}^{(k)}}\nabla J({{\mathbf{\textbf{\emph{P}}}}^{(k)}}) = {\textbf{0}}$, which cannot be chosen as a descent direction. Based on \cite{46}, we find the descent direction as follows.
\par 1) Compute a new vector,
\begin{equation}\label{eq32}
{{\mathbf{\textbf{\emph{q}}}}^{(k)}} = {({\mathbf{\textbf{\emph{N}}}}{{\mathbf{\textbf{\emph{N}}}}^\text{T}})^{ - 1}}{\mathbf{\textbf{\emph{N}}}}\nabla J = {\left( {{{\mathbf{\textbf{\emph{N}}}}^{\text{T}}}} \right)^{ - 1}}\nabla J
\end{equation}
\par 2) Break $\textbf{\emph{q}}^{(k)}$ into two parts ${\mathbf{\textbf{\emph{q}}}}_1^{(k)}$  and ${\mathbf{\textbf{\emph{q}}}}_2^{(k)}$, namely,
\begin{equation}\label{eq33}
{{\mathbf{\textbf{\emph{q}}}}^{(k)}} = \left[ \begin{gathered}
  {\mathbf{\textbf{\emph{q}}}}_1^{(k)} \hfill \\
  {\mathbf{\textbf{\emph{q}}}}_2^{(k)} \hfill \\
\end{gathered}  \right] = {\left[ {\begin{array}{*{20}{c}}
  {{{\left( {{\mathbf{\textbf{\emph{q}}}}_1^{(k)}} \right)}^{\text{T}}}}&{{{\left( {{\mathbf{\textbf{\emph{q}}}}_2^{(k)}} \right)}^{\text{T}}}}
\end{array}} \right]^{\text{T}}}
\end{equation}
where the size of ${{\textbf{\emph{q}}}}_1^{(k)}$  is the number of rows of ${\mathbf{\textbf{\emph{A}}}}_1^{(k)}$, and that of ${\mathbf{\textbf{\emph{q}}}}_2^{(k)}$  is the number of rows of \textbf{\emph{E}}.
\par 3) If all elements of ${{\textbf{\emph{q}}}}_1^{(k)}$  are greater than or equal to 0, i.e. ${\mathbf{\textbf{\emph{q}}}}_1^{(k)} \geq {\textbf{0}}$, then stop. Otherwise, choose any one that is less than 0, delete the corresponding row of ${\mathbf{\textbf{\emph{A}}}}_1^{(k)}$, and use (\ref{eq27})-(\ref{eq30}) to compute ${{\mathbf{\textbf{\emph{d}}}}^{(k)}}$.
\begin{figure}[htb]
\centering
\includegraphics[width=0.2\textwidth]{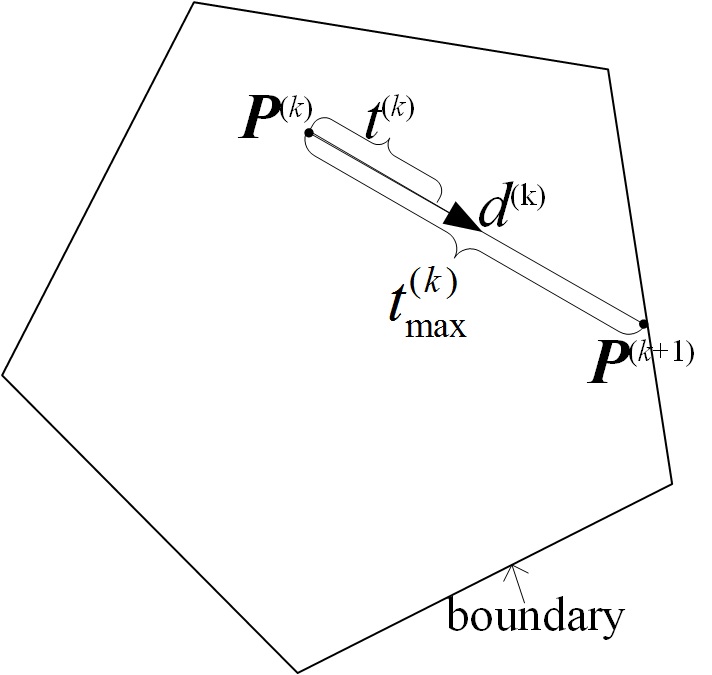}
\caption{A comparison of step length ${t^{{\text{(}}k{\text{)}}}}$  for AGP and $t_{\max }^{{\text{(}}k{\text{)}}}$  for MSAGP.}
\label{fig_2}
\end{figure}

\subsection{The MSAGP  Method}\label{The MSAGP  Method}
In the AGP method, the step length is manually chosen as a small number, leading to slow convergence. To address this issue, we present the MSAGP method by iteratively estimating the maximum step length in the feasible region. At iteration \emph{k}, because ${\mathbf{\textbf{\emph{A}}}}_1^{(k)}{{\mathbf{\textbf{\emph{P}}}}^{(k)}} = {\textbf{0}}$ and ${\mathbf{\textbf{\emph{A}}}}_2^{(k)}{{\mathbf{\textbf{\emph{P}}}}^{(k)}} > {\textbf{0}}$, we can estimate the maximum step length
\begin{equation}\label{eq34}
{t^{{\text{(}}k{\text{)}}}} = t_{\max }^{{\text{(}}k{\text{)}}}
\end{equation}
as follows.
\par 1) Let $p_{ij}^{(k + 1)} = p_{ij}^{(k)} + {t_{ij}}d_{ij}^{(k)} = 0$ for $p_{ij}^{(k)} > 0$,
\par 2) Compute ${t_{ij}} =  - \frac{{p_{ij}^{(k)}}}{{d_{ij}^{(k)}}}$ for $p_{ij}^{(k)} > 0$ and $d_{ij}^{(k)} < 0$,
\par 3) $t_{\max }^{{\text{(}}k{\text{)}}} = \min \left\{ {{t_{ij}}} \right\}$.
\par In Fig. \ref{fig_2}, we can see that step length ${t^{{\text{(}}k{\text{)}}}}$ for AGP is shorter than $t_{\max }^{{\text{(}}k{\text{)}}}$ for MSAGP at iteration \emph{k}. Thus, MSAGP is expected to converge faster than AGP because it always finds a boundary point ${{\mathbf{\textbf{\emph{P}}}}^{{\text{(}}k{\text{ + 1)}}}}$ to update probability vector ${{\mathbf{\textbf{\emph{P}}}}^{{\text{(}}k{\text{)}}}}$. The faster convergence will be confirmed in Section 4.
\subsection{The FMSAGP Method}\label{The FMSAGP Method}
Although MSAGP converges faster than AGP, it has to calculate a large number of matrix inversions. In fact, according to (\ref{eq28}), the calculation of ${{\mathbf{\textbf{\emph{G}}}}^{(k)}} = {{\mathbf{\textbf{\emph{N}}}}^{\text{T}}}{({\mathbf{\textbf{\emph{N}}}}{{\mathbf{\textbf{\emph{N}}}}^\text{T}})^{ - 1}}{\mathbf{\textbf{\emph{N}}}}$ iteratively requires ${({\mathbf{\textbf{\emph{N}}}}{{\mathbf{\textbf{\emph{N}}}}^\texttt{T}})^{ - 1}}$ with ${\mathbf{\textbf{\emph{N}}}} = {{\mathbf{\textbf{\emph{N}}}}^{(k)}}$. To calculate ${{\mathbf{\textbf{\emph{G}}}}^{(k)}}$ more efficiently, it needs the following lemma.
\par \textbf{\emph{Lemma.}} Suppose that $\textbf{\emph{n}}$ is a row vector and ${\left[ {{{\mathbf{\textbf{\emph{n}}}}^{\text{T}}},{{\mathbf{\textbf{\emph{N}}}}^{\text{T}}}} \right]^{\text{T}}}$ is full row rank. If projection matrix ${{\mathbf{\textbf{\emph{G}}}}_{\mathbf{\textbf{\emph{N}}}}} = {{\mathbf{\textbf{\emph{N}}}}^{\text{T}}}{({\mathbf{\textbf{\emph{N}}}}{{\mathbf{\textbf{\emph{N}}}}^\text{T}})^{ - 1}}{\mathbf{\textbf{\emph{N}}}}$ and its orthogonal projection matrix ${{\mathbf{\textbf{\emph{Q}}}}_{\mathbf{\textbf{\emph{N}}}}} = {\mathbf{\textbf{\emph{I}}}} - {{\mathbf{\textbf{\emph{G}}}}_{\mathbf{\emph{\textbf{N}}}}}$, then
\begin{equation}\label{eq35}
	{{\mathbf{\textbf{\emph{G}}}}_{{{\left( {{{\mathbf{\textbf{\emph{n}}}}^{\text{T}}},{{\mathbf{\textbf{\emph{N}}}}^{\text{T}}}} \right)}^{\text{T}}}}} = {{\mathbf{\textbf{\emph{G}}}}_{\mathbf{\textbf{\emph{N}}}}} + {{\mathbf{\textbf{\emph{Q}}}}_{\mathbf{\textbf{\emph{N}}}}}{{\mathbf{\textbf{\emph{n}}}}^{\text{T}}}{\left\langle {{{\mathbf{\textbf{\emph{Q}}}}_{\mathbf{\textbf{\emph{N}}}}}{{\mathbf{\textbf{\emph{n}}}}^{\text{T}}},{{\mathbf{\textbf{\emph{Q}}}}_{\mathbf{\textbf{\emph{N}}}}}{{\mathbf{\textbf{\emph{n}}}}^{\text{T}}}} \right\rangle ^{ - 1}}{\mathbf{\textbf{\emph{n}}}}{{\mathbf{\textbf{\emph{Q}}}}_{\mathbf{\textbf{\emph{N}}}}}.	
\end{equation}	
\par \textbf{Proof.} Because ${\left[ {{{\mathbf{\textbf{\emph{n}}}}^{\text{T}}},{{\mathbf{\textbf{\emph{N}}}}^{\text{T}}}} \right]^{\text{T}}}$ is full row rank, by matrix theory \cite{48}\! -\!\cite{50}, there must exist a matrix \textbf{\emph{G}} such that
\begin{equation}\label{eq36}
	\left\{ \begin{gathered}
  {{\mathbf{\textbf{\emph{G}}}}_{{{\left( {{{\mathbf{\textbf{\emph{n}}}}^{\text{T}}},{{\mathbf{\textbf{\emph{N}}}}^{\text{T}}}} \right)}^{\text{T}}}}} = {\mathbf{\textbf{\emph{G}}}} + {{\mathbf{\textbf{\emph{G}}}}_{\mathbf{\textbf{\emph{N}}}}}, \hfill \\
  {\mathbf{\textbf{\emph{G}}}}{{\mathbf{\textbf{\emph{G}}}}_{\mathbf{\textbf{\emph{N}}}}} = {{\mathbf{\textbf{\emph{G}}}}_{\mathbf{\textbf{\emph{N}}}}}{\mathbf{\textbf{\emph{G}}}} = \textbf{0} \hfill \\
\end{gathered}  \right.,
\end{equation}	
Meanwhile, there must exist a matrix \textbf{\emph{Z}}, such that
\begin{equation}\label{eq37}
	{\mathbf{\textbf{\emph{G}}}} = {{\mathbf{\textbf{\emph{Z}}}}^{\text{T}}}{{\mathbf{\textbf{\emph{n}}}}^{\text{T}}}{\left\langle {{{\mathbf{\textbf{\emph{Z}}}}^{\text{T}}}{{\mathbf{\textbf{\emph{n}}}}^{\text{T}}},{{\mathbf{\textbf{\emph{Z}}}}^{\text{T}}}{{\mathbf{\textbf{\emph{n}}}}^{\text{T}}}} \right\rangle ^{ - 1}}{\mathbf{\textbf{\emph{nZ}}}}	
\end{equation}
Combining (\ref{eq36}) with (\ref{eq37}), we obtain
\begin{equation}\label{eq38}
	{{\mathbf{\textbf{\emph{Z}}}}^{\text{T}}}{{\mathbf{\textbf{\emph{n}}}}^{\text{T}}}{\left\langle {{{\mathbf{\textbf{\emph{Z}}}}^{\text{T}}}{{\mathbf{\textbf{\emph{n}}}}^{\text{T}}},{{\mathbf{\textbf{\emph{Z}}}}^{\text{T}}}{{\mathbf{\textbf{\emph{n}}}}^{\text{T}}}} \right\rangle ^{ - 1}}{\mathbf{\textbf{\emph{nZ}}}}{{\mathbf{\textbf{\emph{G}}}}_{\mathbf{\textbf{\emph{N}}}}} = {\textbf{0}}
\end{equation}
Thus, ${\mathbf{\textbf{\emph{nZ}}}}{{\mathbf{\textbf{\emph{G}}}}_{\mathbf{\textbf{\emph{N}}}}} = {\textbf{0}}$ must hold for all \emph{\textbf{n}}. This leads to ${\mathbf{\textbf{\emph{Z}}}}{{\mathbf{\textbf{\emph{G}}}}_{\mathbf{\emph{\textbf{N}}}}} = {\textbf{0}}$. That is to say, \textbf{\emph{Z}} is the orthogonal projection matrix of ${{\mathbf{\textbf{\emph{G}}}}_{\mathbf{\textbf{\emph{N}}}}}$, ${\mathbf{\textbf{\emph{Z}}}} = {\mathbf{\textbf{\emph{I}}}} - {{\mathbf{\textbf{\emph{G}}}}_{\mathbf{\emph{\textbf{N}}}}} = {{\mathbf{\textbf{\emph{Q}}}}_{\mathbf{\emph{\textbf{N}}}}}$. Moreover, both ${\textbf{\emph{G}}_\textbf{\emph{N}}}$  and  ${\textbf{\emph{Q}}_\textbf{\emph{N}}}$ are  symmetric matrices.
Hence, (\ref{eq35}) holds.

\par Using the above lemma, we can devise the FMSAGP method as follows,
\par 1) Let \emph{k} = 0, ${\mathbf{\textbf{\emph{N}}}} = {{\mathbf{\textbf{\emph{N}}}}^{{\text{(0)}}}} = {\mathbf{\textbf{\emph{E}}}}$, ${{\mathbf{\textbf{\emph{G}}}}^{{\text{(0)}}}} = {{\mathbf{\textbf{\emph{N}}}}^{\text{T}}}{({\mathbf{\textbf{\emph{N}}}}{{\mathbf{\textbf{\emph{N}}}}^\texttt{T}})^{ - 1}}{\mathbf{\textbf{\emph{N}}}}$, initialize $\textbf{\emph{P}}^{(0)}$, go to 4);
\par 2) Find an active row vector \textbf{\emph{n}} from ${\mathbf{\emph{\textbf{A}}}}_2^{(k - 1)}$, such that ${{\mathbf{\textbf{\emph{n}}}}^{\text{T}}}{{\mathbf{\textbf{\emph{P}}}}^{(k - 1)}}{ > \textbf{0}}$, ${{\mathbf{\textbf{\emph{n}}}}^{\text{T}}}{{\mathbf{\textbf{\emph{P}}}}^{(k)}} = {\textbf{0}}$;
\par 3) Compute ${{\mathbf{\textbf{\emph{G}}}}^{(k)}}$ as
\begin{equation}\label{eq39}
	{{\mathbf{\textbf{\emph{G}}}}^{(k)}} = {{\mathbf{\textbf{\emph{G}}}}^{{\text{(}}k{\text{ - 1)}}}} + {{\mathbf{\textbf{\emph{Q}}}}^{{\text{(}}k{\text{ - 1)}}}}{{\mathbf{\textbf{\emph{n}}}}^{\text{T}}}{\left\langle {{{\mathbf{\textbf{\emph{Q}}}}^{{\text{(}}k{\text{ - 1)}}}}{{\mathbf{\emph{\textbf{n}}}}^{\text{T}}},{{\mathbf{\textbf{\emph{Q}}}}^{{\text{(}}k{\text{ - 1)}}}}{{\mathbf{\textbf{\emph{n}}}}^{\text{T}}}} \right\rangle ^{ - 1}}{\mathbf{\textbf{\emph{n}}}}{{\mathbf{\emph{\textbf{Q}}}}^{{\text{(}}k{\text{ - 1)}}}}
\end{equation}
\par 4) Compute ${{\mathbf{\textbf{\emph{Q}}}}^{{\text{(}}k{\text{)}}}} = {{\mathbf{\textbf{\emph{I}}}}_{LK \times LK}} - {{\mathbf{\textbf{\emph{G}}}}^{{\text{(}}k{\text{)}}}}$, and ${{\mathbf{\textbf{\emph{d}}}}^{{\text{(}}k{\text{)}}}} =  - {{\mathbf{\textbf{\emph{Q}}}}^{{\text{(}}k{\text{)}}}}\nabla J({{\mathbf{\textbf{\emph{P}}}}^{{\text{(}}k{\text{)}}}})$;
\par 5) Compute step length $\emph{t}^{(k)}$ by (\ref{eq34}) and probability vector $\textbf{\emph{P}}^{(k+1)}$ by (\ref{eq31});
\par 6) Let \emph{k} = \emph{k} + 1, if ${{\mathbf{\textbf{\emph{G}}}}^{(k)}} \ne {{\mathbf{\textbf{\emph{I}}}}_{LK \times LK}}$, go to 2);
\par 7) Construct ${\mathbf{\textbf{\emph{N}}}} = {{\mathbf{\textbf{\emph{N}}}}^{(k)}}$ by (\ref{eq27}), and compute a new vector ${{\mathbf{\textbf{\emph{q}}}}^{(k)}} = {\left( {{{\mathbf{\textbf{\emph{N}}}}^{\text{T}}}} \right)^{ - 1}}\nabla J$ by (\ref{eq32});
\par 8) Break $\textbf{\emph{q}}^{(k)}$ into two parts ${\mathbf{\textbf{\emph{q}}}}_1^{(k)}$ and ${\mathbf{\textbf{\emph{q}}}}_2^{(k)}$ by (\ref{eq33});
\par 9) If all elements of ${\mathbf{\textbf{\emph{q}}}}_1^{(k)}$ are greater than or equal to 0, stop;
\par 10) Choose any element less than 0 from ${\mathbf{\textbf{\emph{q}}}}_1^{(k)}$, and delete the corresponding row of ${\mathbf{\textbf{\emph{A}}}}_1^{(k)}$;
\par 11) Reconstruct ${\mathbf{\textbf{\emph{N}}}} = {{\mathbf{\textbf{\emph{N}}}}^{(k)}}$ by (\ref{eq27});
\par 12) Compute $\textbf{\emph{d}}^{(k)} $ by (\ref{eq28})-(\ref{eq30}), ${t^{(k)}}$ by (\ref{eq34}), and $\textbf{\emph{P}}^{(k+1)}$ by (\ref{eq31});
\par 13) Let \emph{k} = \emph{k} + 1, and go to 7).

\section{Experiments}\label{Experiments}
In order to evaluate the  performance of PKM and how well the methods of AGP, MSAGP, and FMSAGP solve it, we conduct a large number of experiments on fifteen datasets: one artificial dataset, one human face dataset, and thirteen UCI datasets. These datasets are detailed in Table \ref{table_2}.

\begin{table}[htb]
\renewcommand{\arraystretch}{1.3}
\caption
{Details of Artificial and Benchmark Datasets.}
\label{table_2}
\centering
\begin{tabular}{|c|c|c|c|c|}
\hline
  \multicolumn{2}{|c|}{ }  & Instance&    Class & Dimension\\
\hline
  \multicolumn{2}{|c|}{ Artificial dataset }	&310&	4&	2  \\
\hline
\multicolumn{2}{|c|}{ Yale-faces-B }&5850&10&1200 \\
\hline
 \multirow{9}*{UCI}&Iris&150&3&4\\
 \cline{2-5}
 ~&   Parkinson	&195	&2	&22 \\
  \cline{2-5}
 ~& Seeds&	210&	3&	7           \\
  \cline{2-5}
 ~& Segmentation&	210&	7&	19             \\
   \cline{2-5}
  ~& Glass&	214&	6&	9 \\
     \cline{2-5}
  ~&   Ionosphere	&351&	2	&33             \\
     \cline{2-5}
  ~& Dermatology&	358&	6&	34       \\
     \cline{2-5}
  ~&  Breast-cancer&	683&	2&	9       \\
     \cline{2-5}
  ~&  Natural&	2000&	9	&294         \\
  \cline{2-5}
  ~&  Yeast&	2426&	3	&24     \\
    \cline{2-5}
  ~&Waveform&	5000&	3&	21       \\
    \cline{2-5}
  ~&  Satellite	&6435&	6	&36     \\
    \cline{2-5}
  ~& Epileptic&	11500&	5&	178      \\
  \hline
\end{tabular}
\end{table}

\begin{table*}[htb]
\renewcommand{\arraystretch}{1.3}
\caption{Initialization robustness of PKM, KM++ and FCM on the artificial dataset.}
\label{table_3}
\centering
\begin{tabular}{|c|c|c|c|c|c|c|c|c|c|c|c|}
\hline
\multirow{2}*{ } &\multirow{2}*{ PKM}&\multirow{2}*{KM++ }&\multicolumn{9}{|c|}{	FCM (\emph{m} = 1.05, 1.09, 1.1, 1.3, 1.4, 1.5, 1.9, 2.0, 2.1)}\\
\cline{4-12}
~& ~&~&1.05&	1.09&	1.1&	1.3&	1.4&	1.5	&1.9&	2.0&	2.1\\
\hline
Correct times&	954&  645	&  ---& 687	&693&	780&	691&	0&	0&	0&	0\\
\hline
Correct percentage&	95.4\%&	64.5\%&---	&	68.7\%&	69.3\%&	78.0\%&	69.1\%&	0&	0	&0	&0\\
\hline
\end{tabular}
\end{table*}

\begin{figure}[htb]
\centering
\includegraphics[width=0.3\textwidth]{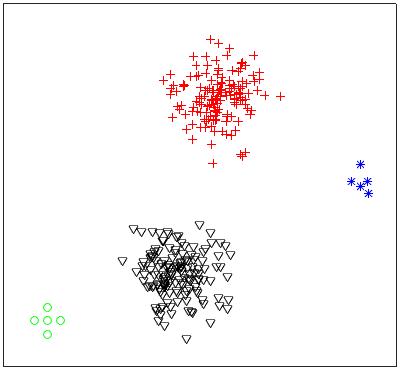}
\caption{Original distribution of the artificial dataset with 4 classes.}
\label{fig_3}
\end{figure}

\begin{figure}[htb]
\centering
\includegraphics[width=0.15\textwidth]{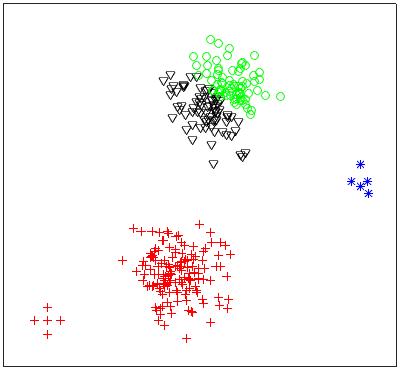}
\includegraphics[width=0.15\textwidth]{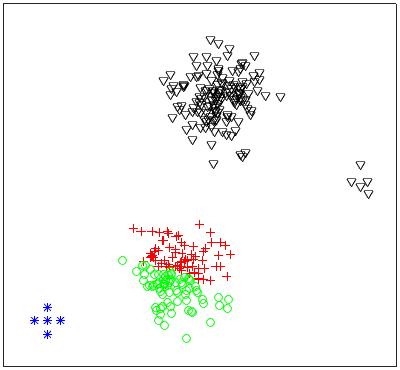}
\includegraphics[width=0.15\textwidth]{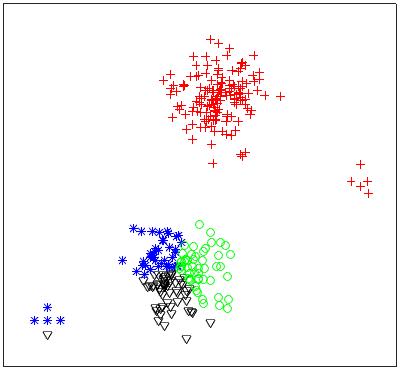} \\
(a)\texttt{ }\texttt{ }\texttt{ }\texttt{ }\texttt{ }\texttt{ }\texttt{ }\texttt{ }\texttt{ }\texttt{ }\texttt{ }\texttt{ }(b)\texttt{ }\texttt{ }\texttt{ }\texttt{ }\texttt{ }\texttt{ } \texttt{ }\texttt{ }\texttt{ }\texttt{ }\texttt{ }\texttt{ }(c)
\caption{Three inconsistent distributions with the artificial dataset.}
\label{fig_4}
\end{figure}

\par The artificial dataset is generated by ourselves. As shown in Fig. \ref{fig_3}, it is composed of 310 data points, with 150, 150, 5 and 5 respectively from 4 classes. The human face dataset is Yale-faces-B\footnote{https://cervisia.org/machine\_learning\_data.php}, which consists of 5850 images scaled down to $30\times40$ pixels, with each saved as a 1200-dimensional vector. The thirteen UCI datasets are selected from UCI Machine Learning Repository\footnote{http://archive.ics.uci.edu/ml}, including Iris, Parkinson, Seeds, Segmentation, Glass, Ionosphere, Dermatology, Breast-cancer, Natural, Yeast, Waveform, Satellite, and Epileptic.
\par All experiments are carried out on the same computer (i7-4790 CPU, 3.60GHz, 8.00G RAM), running Windows 7 and Matlab 2015a. The results of FCM are obtained by Matlab's build-in function "fcm". The results of K-means++ (KM++) are obtained by Matlab's build-in function "kmeans". The results of PKM are obtained by Matlab implementation of AGP, MSAGP and FMSAGP, using two build-in functions "sparse" and "full" for matrix optimization.
\par The experiments are organized and analyzed in five aspects: initialization robustness, clustering performance, descent stability, iteration number, and convergence speed. The first two respects are used to evaluate the performance of PKM, and the last three respects are used to evaluate how well the methods of AGP, MSAGP, and FMSAGP solve  PKM.
\subsection{Initialization Robustness}\label{Initialization}
In this subsection, we evaluate initialization robustness of PKM, KM++ and FCM on the artificial dataset, with the result given in Table \ref{table_3}. The initialization robustness is defined as the number of correct times out of running an algorithm 1000 times initialized randomly. The correctness means the cluster assignation of a dataset is completely consistent with its original distribution. With the artificial dataset in Fig. \ref{fig_3}, three inconsistent distributions are displayed as examples in Fig. \ref{fig_4}. From Table \ref{table_3}, we can see that PKM have 954 correct times, more than 645 by KM++. However, FCM may have number of correct times that varies with parameter \emph{m}. For example, it has 687, 693, 780 and 691 correct times  at \emph{m} = 1.09, 1.1, 1.3 and 1.4, respectively. However, it has no correct times at \emph{m} = 1.5, 1.09, 2.0 and 2.1. At \emph{m} = 1.3, FCM achieves the maximum number of correct times, namely 780, still lower than PKM.
\par Overall, PKM has a correct percentage of 95.4\%, better than KM++ (64.5\%) and FCM (0 to 78.0\%) in terms of initialization robustness.

 \subsection{Clustering Performance}\label{Clustering}
In this subsection, we evaluate clustering performance of PKM, KM++ and FCM in terms of five measures: Standard Squared Error (SSE), Davies Bouldin Index (DBI) \cite{51}, Normalized Mutual Information (NMI) \cite{52}, Adjusted Rand Index (ARI) \cite{52}, and V-measure (VM) \cite{53}. The SSE is defined as
\begin{equation}\label{eq49}
SSE = \sum\limits_{i = 1}^K {\sum\limits_{{{\mathbf{\emph{\textbf{x}}}}_i} \in {\omega _j}} {{{\left\| {{{\mathbf{\textbf{\emph{x}}}}_i} - {{\mathbf{\emph{\textbf{c}}}}_j}} \right\|}^2}} }
\end{equation}
where \emph{K} is the number of clusters, and $\textbf{\emph{c}}_{j}$ is the center (mean) of the \emph{j}-th cluster. DBI is the ratio of within-class distances to between-class distances. NMI is a normalization of the mutual information score to scale the results between 0 (no mutual information) and 1 (perfect correlation). ARI is an adjusted similarity measure between two clusters. V-measure is a harmonic mean between homogeneity and completeness, where homogeneity means that each cluster is a subset of a single class, and completeness means that each class is a subset of a single cluster.
\begin{table*}[htb]
\renewcommand{\arraystretch}{1.3}
\caption{Internal Evaluation of PKM, KM++ and FCM ( \emph{m} = 1.3).}
\label{table_4}
\centering
\begin{tabular}{|c|c|c|c|c|c|c|}
\hline
\multirow{2}*{Datasets} &\multicolumn{3}{|c|}{SSE}&\multicolumn{3}{|c|}{DBI}\\
\cline{2-7}
~& PKM&	KM++&	FCM&	PKM&	KM++&	FCM\\
\hline
Yale-faces-B&	8.51e9  &	8.52e9&	\textbf{8.42e9}&	1.6184&	1.6128&	\textbf{1.5986}\\
\hline
Iris&	\textbf{78.942}&	110.89&	110.9&	\textbf{0.3928}&	0.4801&	0.4561\\
\hline
Seeds&	\textbf{587.05}&	588.05&	587.32&	0.3980&	0.4421&	\textbf{0.3962}\\
\hline
Breast-cancer&	\textbf{2419.3}&	\textbf{2419.3}&	\textbf{2419.3}&	\textbf{0.3785}&	0.3786&	0.3786\\
\hline
Dermatology&	\textbf{11702}&	\textbf{11702}&	\textbf{11702}&	\textbf{0.7180}&	0.7939&	0.7609\\
\hline
Ionosphere&	\textbf{2419.4}&	\textbf{2419.4}&	\textbf{2419.4}&	\textbf{0.7567}&	\textbf{0.7567}&	0.7579\\
\hline
Parkinson&	\textbf{1.34e6}&	\textbf{1.34e6}&	\textbf{1.34e6}	&\textbf{0.4932}&	0.4974&	0.4974\\
\hline
Glass&	372.77&	379.41&	\textbf{366.35}&	0.8505&	\textbf{0.7074}&	0.7909\\
\hline
Natural&	1888.3&	1883.5&\textbf{	1879.2}&	1.7035&	1.6587&	\textbf{1.6089}\\
\hline
Satellite&	\textbf{1.67e7}	&1.69e7	&1.69e7&	\textbf{0.8524}&	0.8731&	0.8957\\
\hline
Yeast&	713.89&	\textbf{713.88}&	714.07&	1.1601&	\textbf{1.1582}&	1.1753\\
\hline
Segmentation&	1.01e6&	\textbf{9.85e5}&	1.99e6&	\textbf{0.7461}&	0.7530&	1.0899\\
\hline
Waveform&	\textbf{1.33e5}&	\textbf{1.33e5}&	\textbf{1.33e5}&	\textbf{0.8663}&	0.9301&	0.8665\\
\hline
Epileptic&	\textbf{5.1e10}&	\textbf{5.1e10}&	5.3e10&	\textbf{2.6051}&	2.6956&	3.6407\\
\hline
\end{tabular}
\end{table*}

\par The above five measures fall into two types: internal evaluation and external evaluation. Internal evaluation, including SSE and DBI, can work without ground truth labels. But external evaluation, such as NMI, ARI and VM, must work with ground truth labels. The lower the internal evaluation, the better the clustering result. However, the higher the external evaluation, the better the clustering result.
\par We test PKM (solved by FMSAGP), KM++ and FCM (\emph{m} = 1.3) on Yale-faces-B and 13 UCI datasets, with the 5-time average results reported in Table \ref{table_4}. The best in each case is highlighted in bold.
Some big values are in scientific notation. For instance, 8.42e9 means $8.42\times10^{9}$.
\par From Table \ref{table_4}, we have the following observations.
\par 1) By SSE, PKM performs better than KM++ on 5 datasets, and as well as KM++ on 6 datasets. Meanwhile, it performs better than FCM on 6 datasets, and as well as FCM on 5 datasets.
\par 2) By DBI, PKM performs better than KM++ on 9 datasets, and as well as KM++ on 1 datasets. Meanwhile, it performs better than FCM on 10 datasets.
\begin{table*}[htb]
\renewcommand{\arraystretch}{1.3}
\caption{External Evaluation of PKM, KM++ and FCM (\emph{m} = 1.3).}
\label{table_5}
\centering
\begin{tabular}{|c|c|c|c|c|c|c|c|c|c|}
\hline
\multirow{2}*{Datasets} &\multicolumn{3}{|c|}{NMI} & \multicolumn{3}{|c|}{ARI} & \multicolumn{3}{|c|}{VM}\\
\cline{2-10}
~& PKM&	KM++&	FCM&	PKM&	KM++&	FCM&	PKM&	KM++&	FCM\\
 \hline
 Yale-faces-B & \textbf{0.7959} & 0.7952 & 0.7722 & \textbf{0.6631} & 0.6123 & 0.6386 & \textbf{0.8264} & 0.1744 & 0.0059 \\
  \hline
 Iris & \textbf{0.7501} & 0.6733 & 0.6723 & \textbf{0.7233} & 0.5779 & 0.5763 & \textbf{0.7419} & 0.7149 & 0.7081 \\
  \hline
 Seeds & \textbf{0.7025} & \textbf{0.7025} & 0.6949 & 0.7135 & 0.7135 & \textbf{0.7166} & \textbf{0.7050} & 0.6999 & 0.6949 \\
  \hline
 Breast-cancer & \textbf{0.7546} & 0.7478 & 0.7478 & \textbf{0.8521} & 0.8464 & 0.8464 & \textbf{0.7545} & 0.7478 & 0.7478 \\
  \hline
 Dermatology & \textbf{0.1141} & 0.1032 & 0.1046 & \textbf{0.0391} & 0.0266 & 0.0261 & \textbf{0.1140} & 0.1056 & 0.1095 \\
  \hline
 Ionosphere & \textbf{0.1349} &\textbf{ 0.1349} & 0.1299 & \textbf{0.1777} & \textbf{0.1777} & 0.1727 & \textbf{0.1348} & \textbf{0.1348} & 0.1298 \\
  \hline
 Parkinson & 0.1124 & \textbf{0.1217} & \textbf{0.1217} & 0.1876 & \textbf{0.2046} & \textbf{0.2046} & 0.1153 & 0.0607 & \textbf{0.1262} \\
  \hline
 Glass & 0.3294 & \textbf{0.4178} & 0.3489 & 0.2201 & \textbf{0.2551} & 0.2126 & \textbf{0.3871} & 0.3857 & 0.3807 \\
  \hline
 Natural & 0.0523 & \textbf{0.0536} & 0.0521 & 0.0247 & 0.0261 & \textbf{0.0273} & 0.0518 & \textbf{0.0529} & 0.0495 \\
  \hline
 Satellite & \textbf{0.6117} & \textbf{0.6117} & 0.5471 & 0.5292 & \textbf{0.5293} & 0.4289 & 0.5790 & 0.5544 & \textbf{0.6128} \\
  \hline
 Yeast & \textbf{0.0050} & \textbf{0.0050} & 0.0043 & \textbf{0.0117} & \textbf{0.0117} & 0.0109 & \textbf{0.0048} & 0.0045 & 0.0045 \\
  \hline
 Segmentation & \textbf{0.5338} & 0.5132 & 0.4678 & \textbf{0.3823} & 0.3313 & 0.3172 & 0.4996 & 0.5252 & \textbf{0.5729} \\
  \hline
 Waveform & \textbf{0.3622} & \textbf{0.3622} & 0.3606 & \textbf{0.2536} & \textbf{0.2536} & 0.2529 & \textbf{0.3622} & \textbf{0.3622} & 0.3559 \\
  \hline
 Epileptic & 0.1228 & \textbf{0.1675} & 0.0033 & 0.0245 & \textbf{0.0263} & 0.0025 & 0.1226 & \textbf{0.1291} & 0.0058 \\
  \hline

\end{tabular}
\end{table*}

\begin{table*}[htb]
\renewcommand{\arraystretch}{1.3}
\caption{Iteration number for MSAGP and AGP to converge with objective function value on 8 UCI datasets.}
\label{table_6}
\centering
\begin{tabular}{|c|c|c|c|c|c|c|}
\hline
\multirow{2}*{ } &\multicolumn{3}{|c|}{Iteration number}&\multicolumn{3}{|c|}{Objective function value}\\
\cline{2-7}
~& MSAGP&	AGP (\emph{t} = 0.01)&	AGP (\emph{t} = 0.1)&	MSAGP&	AGP (\emph{t} = 0.01)&	AGP (\emph{t} = 0.1)\\
 \hline
 Segmentation & \textbf{1290} & 1338 & 1291 & 986167.4343 & \textbf{986167.4343} & 986167.4343 \\
  \hline
 Parkinson & \textbf{195} & 196 & 196 & \textbf{1343350.02} & \textbf{1343350.02} & \textbf{1343350.02}\\
  \hline
 Dermatology & \textbf{1801} & 6329 & 2247 & 11715.1043 & \textbf{11715.1041} & 11715.1043 \\
  \hline
 Breast-cancer & \textbf{452} & 577 & 462 & \textbf{19323.2} & \textbf{19323.2} & \textbf{19323.2} \\
  \hline
 Seeds & \textbf{421} & 634 & 534 & \textbf{587.32} & \textbf{587.32} & \textbf{587.32} \\
  \hline
 Glass & \textbf{1079} & 190683 & 115145 & 378.9 & \textbf{377.19} & 381.74 \\
  \hline
 Ionosphere & \textbf{368} & 2070 & 524 & \textbf{2419.3} & \textbf{2419.3} & \textbf{2419.3} \\
  \hline
 Iris & \textbf{308} & 53022 & 5573 & \textbf{78.95} & \textbf{78.95} & \textbf{78.95} \\
  \hline
\end{tabular}
\end{table*}

\par From Table \ref{table_5}, we have the following observations.
\par 1) By NMI, PKM performs better than KM++ on 5 datasets, and as well as KM++ on 5 datasets. Meanwhile, it outperforms FCM on 12 datasets.
\par 2) By ARI, PKM perform better than KM++ on 5 datasets, and as well as KM++ on 4 datasets. Meanwhile, it outperforms FCM on 11 datasets.
\par 3) By VM, PKM perform better than KM++ on 9 datasets, and as well as KM++ on 2 datasets. Meanwhile, it outperforms FCM on 11 datasets.
\par Overall, PKM outperforms KM++ and FCM in terms of SSE, DBI, NMI, ARI and VM.

 \subsection{Descent Stability of MSAGP}\label{Descent}
In this subsection, we evaluate the descent stability of MSAGP. Theoretically, AGP needs a small step length to make the value of its objective function stably descend at each iteration and gradually converge to a local minimum. If the step length is too large, the value may descend with oscillation, and even does not converge at all. MSAGP iteratively estimates a maximum step length for AGP in the feasible region of PKM so as to speed up its convergence with fewer iterations. However, will this estimate have any serious influence of oscillation on the convergence of AGP?
\par
\begin{figure}[htb]
\centering
\includegraphics[width=0.15\textwidth]{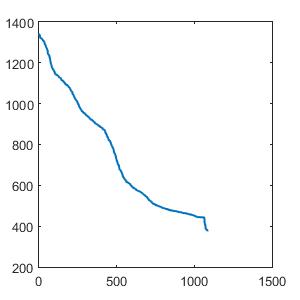}
\includegraphics[width=0.15\textwidth]{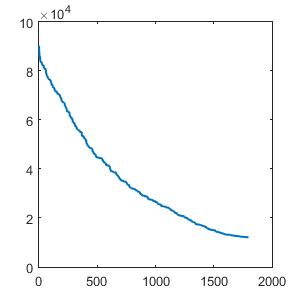}
\includegraphics[width=0.15\textwidth]{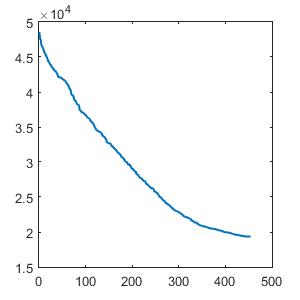} \\
(a)\texttt{ }\texttt{ }\texttt{ }\texttt{ }\texttt{ }\texttt{ }\texttt{ }\texttt{ }\texttt{ }\texttt{ }\texttt{ }\texttt{ }(b)\texttt{ }\texttt{ }\texttt{ }\texttt{ }\texttt{ }\texttt{ } \texttt{ }\texttt{ }\texttt{ }\texttt{ }\texttt{ }\texttt{ }(c)\\

\includegraphics[width=0.15\textwidth]{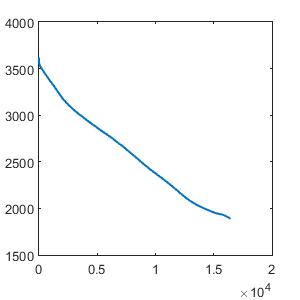}
\includegraphics[width=0.15\textwidth]{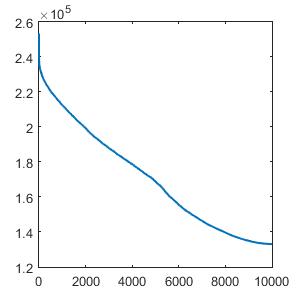}
\includegraphics[width=0.15\textwidth]{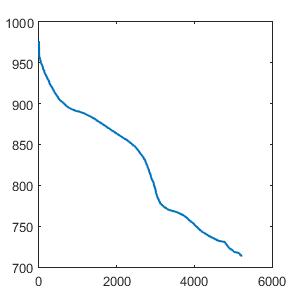} \\
(d)\texttt{ }\texttt{ }\texttt{ }\texttt{ }\texttt{ }\texttt{ }\texttt{ }\texttt{ }\texttt{ }\texttt{ }\texttt{ }\texttt{ }(e)\texttt{ }\texttt{ }\texttt{ }\texttt{ }\texttt{ }\texttt{ } \texttt{ }\texttt{ }\texttt{ }\texttt{ }\texttt{ }\texttt{ }(f)\\

\includegraphics[width=0.15\textwidth]{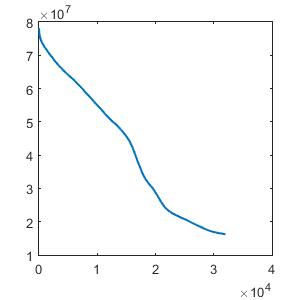}
\includegraphics[width=0.15\textwidth]{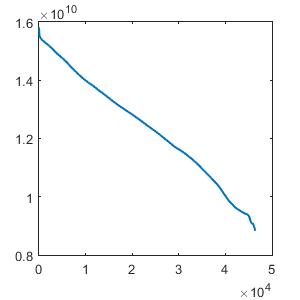}
\includegraphics[width=0.15\textwidth]{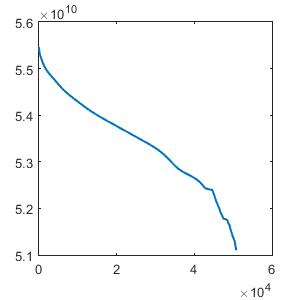} \\
(g)\texttt{ }\texttt{ }\texttt{ }\texttt{ }\texttt{ }\texttt{ }\texttt{ }\texttt{ }\texttt{ }\texttt{ }\texttt{ }\texttt{ }(h)\texttt{ }\texttt{ }\texttt{ }\texttt{ }\texttt{ }\texttt{ } \texttt{ }\texttt{ }\texttt{ }\texttt{ }\texttt{ }\texttt{ }(i)
\caption{Descent stability of MSAGP on 9 UCI datasets.(a) Glass.  (b) Dermatology.       (c) Breast-cancer.
(d) Natural.     (e) Waveform.    f) Yeast. (g) Satellite.      (h) Yale-faces. (i) Epileptic.}
\label{fig_5}
\end{figure}

 To demonstrate the influence, we select 9 UCI datasets from Table \ref{table_2}, including Satellite, Yale, Epileptic, Natural, Yeast, Waveform, Glass, Dermatology, and Breast-cancer. On these datasets, we use MSAGP to solve PKM, and illustrate the value of its objective function vs. iteration in Fig. \ref{fig_5}. Obviously, we can see that MSAGP descends stably without oscillation.

\subsection{Iteration Number of MSAGP and AGP}\label{Iteration}
\begin{figure}[htb]
\centering
\includegraphics[width=0.15\textwidth]{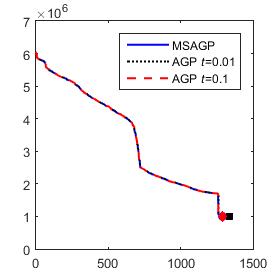}
\includegraphics[width=0.15\textwidth]{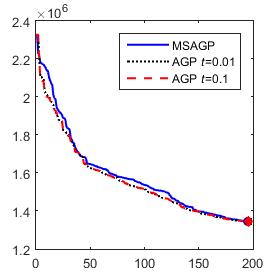}
\includegraphics[width=0.15\textwidth]{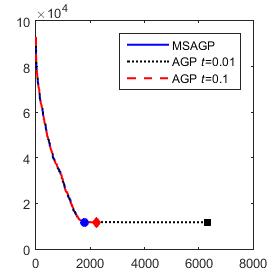} \\
(a)\texttt{ }\texttt{ }\texttt{ }\texttt{ }\texttt{ }\texttt{ }\texttt{ }\texttt{ }\texttt{ }\texttt{ }\texttt{ }\texttt{ }(b)\texttt{ }\texttt{ }\texttt{ }\texttt{ }\texttt{ }\texttt{ } \texttt{ }\texttt{ }\texttt{ }\texttt{ }\texttt{ }\texttt{ }(c)\\

\includegraphics[width=0.15\textwidth]{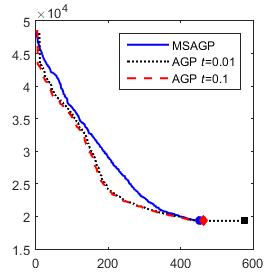}
\includegraphics[width=0.15\textwidth]{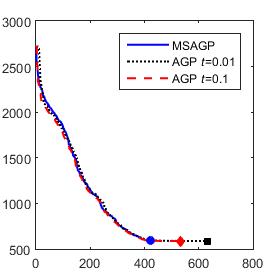}
\includegraphics[width=0.15\textwidth]{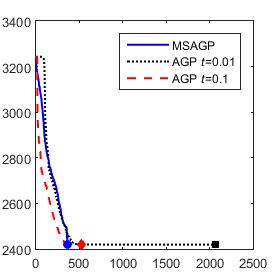} \\
(d)\texttt{ }\texttt{ }\texttt{ }\texttt{ }\texttt{ }\texttt{ }\texttt{ }\texttt{ }\texttt{ }\texttt{ }\texttt{ }\texttt{ }(e)\texttt{ }\texttt{ }\texttt{ }\texttt{ }\texttt{ }\texttt{ } \texttt{ }\texttt{ }\texttt{ }\texttt{ }\texttt{ }\texttt{ }(f)
\caption{Iteration number (in x-axis) for MSAGP and AGP (\emph{t} = 0.01 and 0.1) to converge on 6 UCI datasets: Segmentation (a), Parkinson (b), Dermatology (c), Breast-cancer (d), Seeds (e) and Ionosphere (f). Y-axis means objective function value of PKM.}
\label{fig_6}
\end{figure}

\begin{figure}[htb]
\centering
\includegraphics[width=0.15\textwidth]{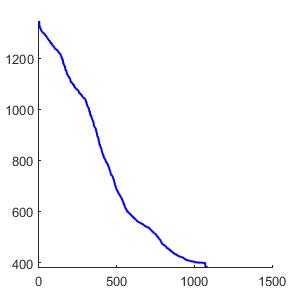}
\includegraphics[width=0.15\textwidth]{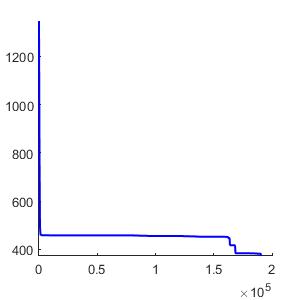}
\includegraphics[width=0.15\textwidth]{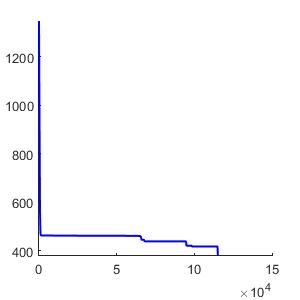} \\
(a)\texttt{ }\texttt{ }\texttt{ }\texttt{ }\texttt{ }\texttt{ }\texttt{ }\texttt{ }\texttt{ }\texttt{ }\texttt{ }\texttt{ }(b)\texttt{ }\texttt{ }\texttt{ }\texttt{ }\texttt{ }\texttt{ } \texttt{ }\texttt{ }\texttt{ }\texttt{ }\texttt{ }\texttt{ }(c)\\

\includegraphics[width=0.15\textwidth]{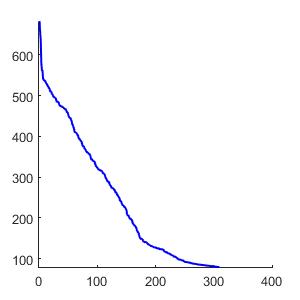}
\includegraphics[width=0.15\textwidth]{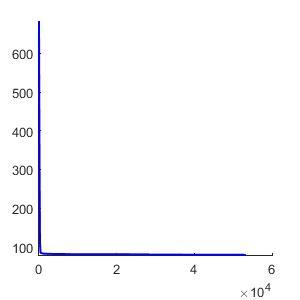}
\includegraphics[width=0.15\textwidth]{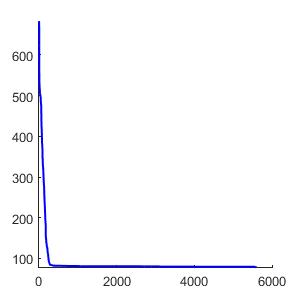} \\
(d)\texttt{ }\texttt{ }\texttt{ }\texttt{ }\texttt{ }\texttt{ }\texttt{ }\texttt{ }\texttt{ }\texttt{ }\texttt{ }\texttt{ }(e)\texttt{ }\texttt{ }\texttt{ }\texttt{ }\texttt{ }\texttt{ } \texttt{ }\texttt{ }\texttt{ }\texttt{ }\texttt{ }\texttt{ }(f)
\caption{Iteration number (in x-axis) for MSAGP (a) and AGP (\emph{t} = 0.01) (b) AGP (\emph{t} = 0.1) (c) to converge on Glass, and for MSAGP (d) and AGP (\emph{t} = 0.01) (e) AGP (\emph{t} = 0.1) (f) on Iris. Y-axis means objective function value of PKM.}
\label{fig_7}
\end{figure}

In this subsection, we compare iteration number for MSAGP and AGP (\emph{t} = 0.01 and 0.1) to converge on 8 UCI datasets, including Segmentation, Parkinson, Dermatology, Breast-cancer, Seeds, Ionosphere, Glass and Iris. The results are illustrated in Figs. \ref{fig_6}-\ref{fig_7} and reported in Table \ref{table_6}, where we have the following observations.
\par 1) MSAGP requires fewer iterations to converge than AGP on the eight datasets, especially much fewer on Dermatology (Fig. \ref{fig_6}c), Ionosphere (Fig. \ref{fig_6}f), Glass (Figs. \ref{fig_7}a-c) and Iris (Figs. \ref{fig_7}d-f).
\par 2) When converging, MSAGP and AGP have the same objective function value on Parkinson, Breast-cancer, Seeds, Ionosphere and Iris. However, MSAGP has an objective function value slightly greater than AGP (\emph{t} = 0.01) on Segmentation, Dermatology and Glass, while slightly smaller than AGP (\emph{t} = 0.1) on Glass.
\par Overall, MSAGP can reach a competitive convergence with AGP in fewer iterations.

\subsection{Convergence Speed of FMSAGP and MSAGP}\label{Convergence}
In this subsection, we compare convergence speed of FMSAGP and MSAGP in running time on 9 UCI datasets: Parkinson, Iris, Seeds, Glass, Segmentation, Breast-cancer, Dermatology, Yeast and Waveform. The results are reported in Table \ref{table_7}, where we have the following observations.
\par 1) FMSAGP runs faster than MSAGP on all the nine datasets.
\par 2) The larger the dataset, the lower the speed-up ($\eta$) of FMSAGP to MSAGP. For example, the speed-up is 47.3\% on Iris (150 samples), 22.3\% on Glass (214 samples), 9.7\% on Yeast (2426 samples) and 1.5\% on Waveform (5000 samples).

\begin{table}[htb]
\renewcommand{\arraystretch}{1.3}
\caption{Running Time (s) of FMSAGP and MSAGP with speed-up on 9 UCI datasets.}
\label{table_7}
\centering
\begin{tabular}{|c|c|c|c|}
\hline
	&FMSAGP &	MSAGP& $\eta$\\
\hline
 Parkinson & 0.1395 & 0.2895 & 48.2\% \\
  \hline
 Iris & 0.2691 & 0.5692 & 47.3\% \\
  \hline
 Seeds & 0.4868 & 1.3117 & 37.1\% \\
  \hline
 Glass & 2.4432 & 10.9369 & 22.3\% \\
  \hline
 Segmentation & 2.7152 & 14.6201 & 18.6\% \\
  \hline
 Breast-cancer & 0.8911 & 5.4898 & 16.2\% \\
  \hline
 Dermatology & 6.1486 & 40.7612 & 15.1\% \\
  \hline
 Yeast & 137.98 & 1418.47 & 9.7\% \\
  \hline
 Waveform & 173.49 & 11957.21 & 1.5\% \\
  \hline

\end{tabular}
\end{table}

\section{Conclusion}\label{Conclusion}
In this paper, the most important contribution is our solution to the clustering problem of FCM at \emph{m} = 1 by a new model, i.e. PKM. Moreover, we have addressed the PKM model by three methods: AGP, MSAGP and FMSAGP. Additionally, we have conducted a large number of experiments to evaluate how well these methods work in initialization robustness, clustering performance, descent stability, iteration number, and convergence speed. As future work, we will further improve our solution to PKM, and apply the basic idea to other relevant problems, like $L_{p}$-norm K-means, kernel PKM, and even possibilistic c-means. Particularly, we will combine deep neural networks \cite{54}, \cite{55} with PKM to build deep PKM models, and to develop more nonlinear programming models for machine learning.


%

%

\ifCLASSOPTIONcompsoc
  \section*{Acknowledgments}
\else
  \section*{Acknowledgment}
\fi

This work was supported by the National Natural Science Foundation of China under grant numbers: 61876010 and 61806013.

\ifCLASSOPTIONcaptionsoff
  \newpage
\fi

\end{document}